# Chaurah: A Smart Raspberry Pi based Parking System

Soumya Ranjan Choudhaury[1], Aditya Narendra[1], Ashutosh Mishra[2] and Ipsit Misra[1]

[1] Center of Excellence-Artificial Intelligence,
Odisha University of Technology and Research, Bhubaneswar, India
`adinarendra0108@gmail.com`
[2] Odisha University of Technology and Research, Bhubaneswar, India

**Abstract** The widespread usage of cars and other large, heavy vehicles necessitates the development of an effective parking infrastructure. Additionally, algorithms for detection and recognition of number plates are often used to identify automobiles all around the world where standardized plate sizes and fonts are enforced, making recognition a effortless task. As a result, both kinds of data can be combined to develop an intelligent parking system focuses on the technology of Automatic Number Plate Recognition (ANPR). Retrieving characters from an inputted number plate image is the sole purpose of ANPR which is a costly procedure. In this article, we propose Chaurah, a minimal cost ANPR system that relies on a Raspberry Pi 3 that was specifically created for parking facilities. The system employs a dual-stage methodology, with the first stage being an ANPR system which makes use of two convolutional neural networks (CNNs). The primary locates and recognises license plates from a vehicle image, while the secondary performs Optical Character Recognition (OCR) to identify individualized numbers from the number plate. An application built with Flutter and Firebase for database administration and license plate record comparison makes up the second component of the overall solution. The application also acts as an user-interface for the billing mechanism based on parking time duration resulting in an all-encompassing software deployment of the study.

## 1  Introduction

The rising number of vehicles on the road on a daily basis, along with the restricted amount of available parking space, necessitates the deployment of multilevel parking systems. These parking systems usually necessitate considerable financial investments in order to function properly. The introduction of AI has triggered a tremendous technological revolution. Basically, all manual and repetitive tasks can be substituted by cognitive computing, leading to the creation of a diverse range of systems, many of which are intelligent decision-making systems, intelligent vision-based systems, and so on. When this is integrated with IoT, one might expect small, effective, and low-cost solutions. Combining AI-based algorithms and IOT devices allows for the creation of systems that are small yet powerful enough to be used in a range of contexts, including hospitals[1], agricultural activities[2], and industrial settings[3]. This paper introduces Chaurah, a Smart Raspberry Pi-based parking system that provides an end-to-end solution for parking systems by utilising AI-based algorithms and a low-cost microcontroller. This comprises ANPR and OCR modules that use CNNs and are subsequently sent to a front end application for use in user-facing modules. Overall, in public parking facilities, this method is economical, efficient, and simple to use.

## 2      Previous Works

The primary focus of the meticulously designed investigation was the research papers which were compiled in Web of Science (SCI/SSCI), one of the most extensively utilised web-based databases. Owing to the fact that Web of Science collects published materials from Science Citation Index (SCI) and the Social Science Citation Index (SSCI). For this investigation, it was chosen as the primary source database. The common keywords that were searched for to collect papers included "ANPR System", "Smart Parking System", "Raspberry Pi based Parking System" etc. These searches returned a total of 128 papers out of which 12 were duplicated and 17 had unclear methodologies in it. The remaining studies underwent an exhaustive literature analysis in the topic, which were inclusive of a study of alternative approaches for Automatic Number Plate Recognition (ANPR) and Optical Character Recognition (OCR).

Based on previous research, it was discovered that the typical approach to the problem consisted of two modules: (a) an ANPR module using a CNN model or object detection models such as YOLO and its variants, and (b) an OCR module using image filtering techniques and then passing through a CNN Network. Several Works[4] [5] [6] have used Raspberry-Pi to achieve these ideas but have used comparable approaches in the ANPR and OCR modules.Table 1 given below lists some of the prior works that used these methodologies.

Table 1. An Overview of Previous works and their Approaches.

| Name of Work | Approach for ANPR | Approach for OCR |
| --- | --- | --- |
| Automatic number plate detection and recognition [7] | YOLO [8] | Otsu's Algorithm [9] + OpenCV |
| Automatic Number Plate Recognition using YOLO for Indian Conditions [10] | Custom CNN | LSTM Tesseract |
| Automatic Number Plate Recognition and Parking Management [11] | YOLOv4 [12] | Custom CNN |
| Automatic License Plate Recognition for Parking System using CNNs [13] | YOLO [8] | ResNet [14] |
| A Robust Real-Time Automatic License Plate Recognition Based on the YOLO Detector [15] | YOLO [8] | Custom CNN |
| An automated vehicle parking monitoring and management system using ANPR cameras [16] | Custom CNN | Custom CNN |

## 3      Proposed System

Chaurah, a low-cost, end-to-end autonomous parking system, is presented in this study. The technology is based on a registration plate recognition system that uses deep learning techniques to gather licence plate numbers for cars parked in a parking lot and provides a smooth user experience via a user-facing application. The proposed approach does not necessitate the use of any extra equipment.

A Pi camera serves as the system's input, capturing the entry of a car into the parking lot and sending it to the Automatic Number Plate Recognition (ANPR) module for further handling. This technique performs a number of processes on the image, including segmentation, localisation, orientation, and normalisation, before returning a cropped version of the registration plate. This truncated picture is then submitted to the optical character recognition (OCR) system, which generates a final output from the image of the licence plate. A Raspberry Pi module serves as the system's foundation, sending data to the cloud server and eventually to the application. After that, the collected data is compared to entries in a database kept on a cloud server. When the vehicle's records match, the parking timer starts, and the user is informed by the phone number they provided when registering for the service. The camera collects an image and tells the ANPR and OCR modules when the car left the parking lot. The number is checked again until a match is found, at which point the parking timer is turned off and the total time spent parking is calculated. The software alerts the user, and the total amount is calculated using the current rate schedule and the duration that was parked. The appropriate amount is then deducted automatically from the parking app wallet, and the entire transaction is captured and saved in the database for future use and review.

The latest results demonstrate that the suggested approach accurately recognises and recognizes the majority of car licence plate classes on real-world images. This technology could potentially be used to regulate traffic and provide security. This method could also be used to identify stolen vehicles on the road. The key benefit of this approach is that it does not require any new hardware on automobiles to function. As a consequence, this solution is comprehensive, cost-effective, and adaptable to a variety of parking facilities. Fig. 1. depicts the general operational structure of the suggested solution.

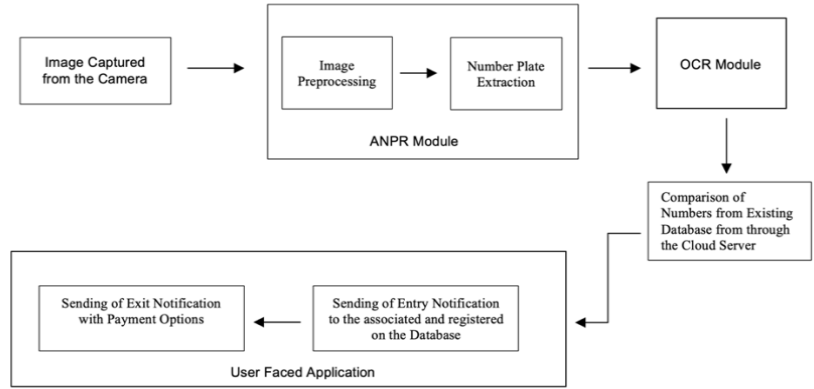

**Fig. 1.** The Overall Working of Chaurah

### 3.1 Hardware Used

The Raspberry Pi 4 serves as the system's hardware, using an ARM v8-based quad-core Cortex-A72 SoC and 4GB of LPDDR4-3200 SDRAM. In addition, we used two Raspberry Pi v2 camera modules, each having an 8-megapixel resolution and capable of shooting videos at 1080p at 30 frames per second.

### 3.2 Dataset Used

Around 4500 photographs of cars taken from various locations, angles, and viewpoints were used to train the ANPR model. The majority of the images were scraped and obtained from Google and Flickr. Over 1200 of these images were discovered in earlier works repositories [17] [18] on Kaggle and GitHub, respectively. Also, 700 pictures were retrieved from another GitHub repository [19]. Fig. 2 shows a selection of photos from the dataset used to develop the ANPR model.

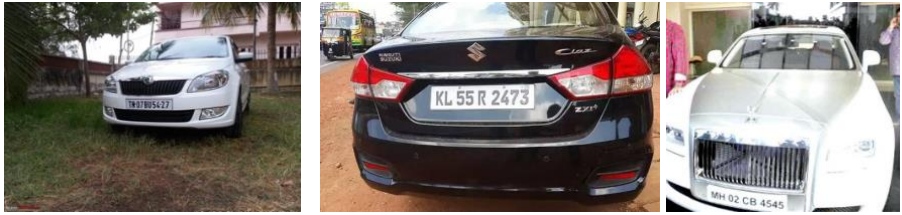

**Fig. 2.** Sample Dataset Images for the ANPR model

## 4 Working and Experiments

The full working of the presented system constitutes three main modules as mentioned below:

4.1. Automatic Number Plate Recognition Module
4.2. The Optical Character Recognition Module
4.3. User Faced Application

The detailed working and experiments for each of the module are mentioned below.

### 4.1 Automatic Number Plate Recognition Module

The identification of a vehicle's license plate is the first stage in the ANPR procedure which utilizes techniques to extract a registration plate's rectangular section from an original image. We decided to evaluate one-stage object identification algorithms to their shorter inference time. Using our dataset, experiments were conducted on various one-stage object detection models for the ANPR module, including Single Shot Detection (SSD) [20], RetinaNet [21], YOLOv4 and YOLOv3 [22]. The details of accuracies are given below in Table 2.

**Table 2.** Training and Testing Accuracies of Various Object Detection Models

| Name of the Model | Training Accuracy (in Percent) | Testing Accuracy (in Percent) | Mean Average Precision | Average Frames Per Second |
|---|---|---|---|---|
| SSD | 92.38 | 91.42 | 74.65 | 4 |
| RetinaNet | 91.46 | 89.38 | 86.83 | 7 |
| **YOLOv4** | **96.22** | **92.65** | **88.25** | **16** |
| YOLOv3 | 90.54 | 89.78 | 84.42 | 14 |

We chose the YOLOv4 model for the proposed system because of its small size, quick processing time, and high accuracy. The YOLO technique was first proposed by Redmon et al. in their work "You Only Look Once: Unified, Real-Time Detection." In addition to its simple architecture, their research demonstrates that YOLO recognizes objects quickly and accurately, with ImageNet 2012 validation reaching 88%. This model was refined into YOLOv2 [23], YOLOv3, and lastly YOLOv4, all of which performed better on benchmarking datasets. YOLOv4 is a one-stage object detection model based on the original YOLO models. Object detection algorithms are typically separated into two parts: the head and the backbone. The backbone is usually previously trained on a larger image classification dataset, like ImageNet, and it works to encode pertinent data about the input. The head predicts object classes and information about bounding boxes. The YOLOv4 paper also describes a "neck," which are layers between the backbone and the head that collect feature mappings from different stages of the network. YOLOv4 is built based upon current studies , utilizing CSPDarknet53 as the Backbone, SPP (Spatial Pyramid Pooling) and PAN (Path Aggregation Network) as the "Neck," and YOLOv3 as the "Head."

The YOLOv4 study collects extra training methodologies and labels them as "bag-of-freebies" or "bag-of-specials" (BoS). Bag-of-Freebies are training approaches that have no effect on the training plan or budget. BoS are training strategies that increase inference costs somewhat while potentially increasing model performance. The overall architecture of the system is algorithm from the original paper is depicted through Fig. 3.

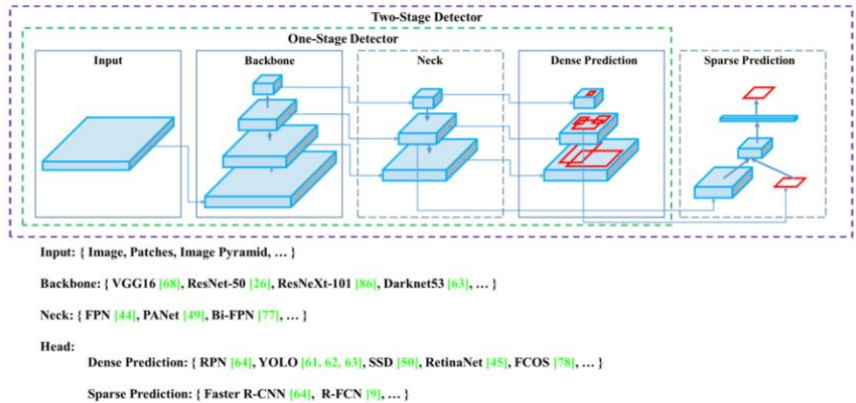

**Fig. 3.** The Architecture of YOLOv4

The collected data consisting of around 4500 images along with their corresponding bounding boxes was utilized for the training of YOLOv4 model. The loss vs epochs graph was produced shown in Fig. 4. Also few sample images for the output generated from the module are shown below in Fig. 5.

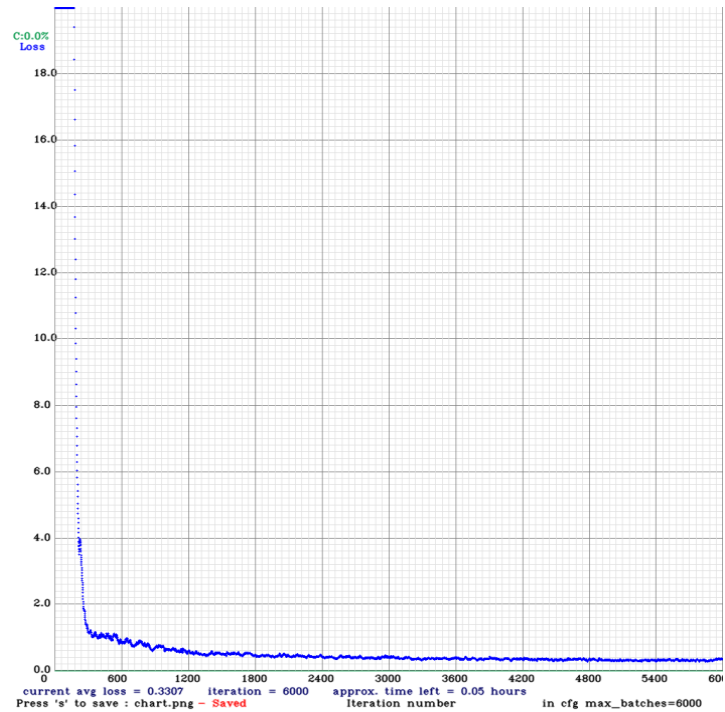

**Fig. 4.** Loss Vs Epochs Graph during Training

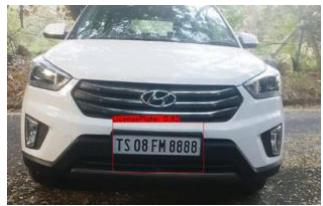
(a)

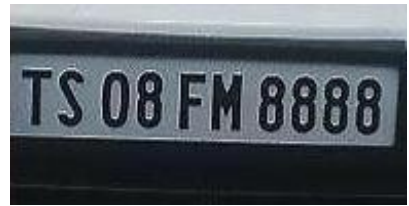
(b)

**Fig. 5.** (a) ANPR Module detecting the number plate (b) Cropped Image from ANPR Module

### 4.2    Optical Character Recognition (OCR) Module

This is the procedure for converting a text image into machine-readable text. There are a variety of OCR models accessible, including past works. Due to its small size, excellent accuracy, and fast-reliable processing, the Keras OCR pipeline is employed in this module for digit extraction from the number plate photo.

The KerasOCR pipeline as a whole offers out-of-the-box OCR models as well as a rigorous training methodology for providing new OCR models. The model includes Convolutional Neural Networks as both a sensor and a recognition systems. The detector's objective, as indicated by its name, is to detect characters in a picture, whereas the recognizer's goal is to accurately identify and return characters to the user as a string. The basic stage taken by this system is to recognise the characteristics in the image and decode it for further interpretation, as shown in Fig. 6. The bounding box method is used to collect individual characters from the licence plate, which are then matched to an alphanumeric database using matching technique. The retrieved character is contrasted with the model pictures at any feasible locations. A huge library of alpha numeric characters is accessible, with different font styles and sizes for each character for comparison. The finished result is presented as a notepad file/displayed terminal which is shown in Fig. 7.

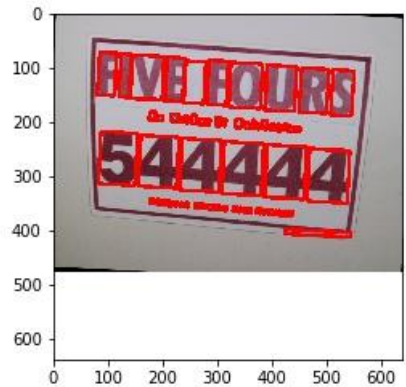

**Fig. 6.** Sample Output from the Detector of OCR Module

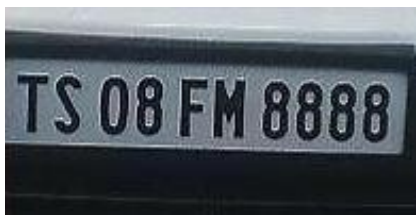

(a)                  (b)

**Fig. 7.** (a) Input Sample from ANPR Module to OCR Module (b) Extracted Number with inference time on Terminal from OCR Module

The retrieved license plate number is then transferred to a cloud-based server and compared to a database of previously registered records. When a match is detected, the parking garage is opened, and the user is notified with a time stamp via the app. Similarly, once the vehicle exits the parking facility, the entire system is triggered once more.

### 4.3 User Facing Application

The application's major purpose was to develop a simple and transparent mechanism to deliver a user-friendly experience. The app was built with Flutter and uses Firebase for database management. The application's essential features are described below.

1. Display the current parking period as well as the overall parking duration.
2. Provide real-time notification of parking facility entry/exit.
3. Provide the user with a parking wallet so that they can smoothly transact their parking trips.
4. Offer the user a hassle free mechanism to add money to their parking wallet.
5. Based on the current rate schedule, determine the total cost of each trip's parking fees
6. Preserve records of any previous parking facility visits, including their duration, timing, cost, and other information, for future use.

Furthermore, this output from the application can be customized to be put on display on a screen or at a control center as per the requirements of the concerned authority.
The interfaces for various functions of the application are shown below in Fig. 8.

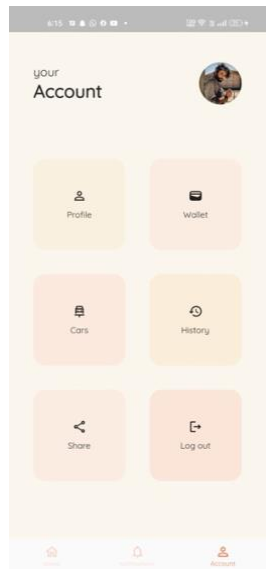
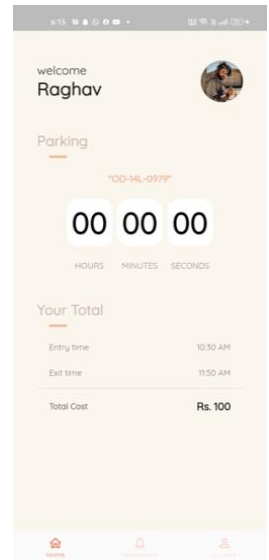

(a)  (b)

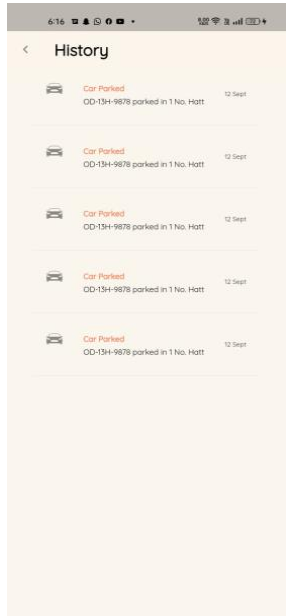
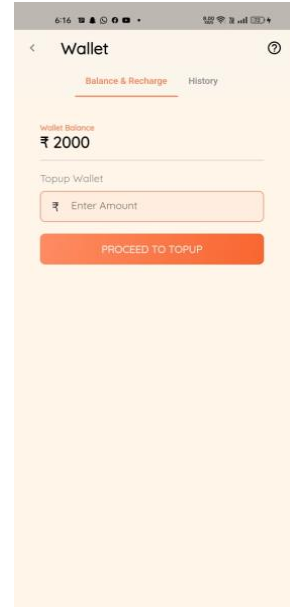

(c)                                                   (d)

**Fig. 8.** (a) Home page of the application (b) Sample Parking Bill generated by the App (c) Log of all past trips (d) Parking Wallet in the App

## 5 Results

The entire system is illustrated in Fig.9. below, along with an example of how the system works from its inception to its conclusion.

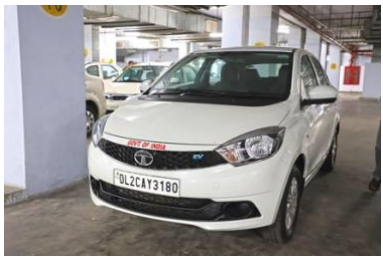
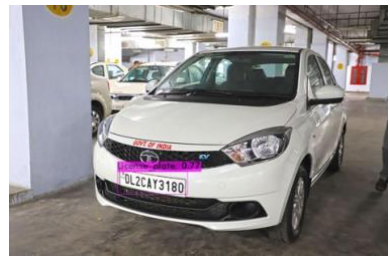

(a)                                                   (b)

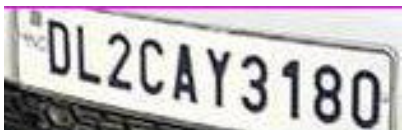
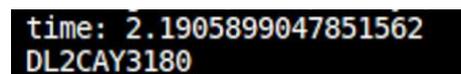

(c)                                                   (d)

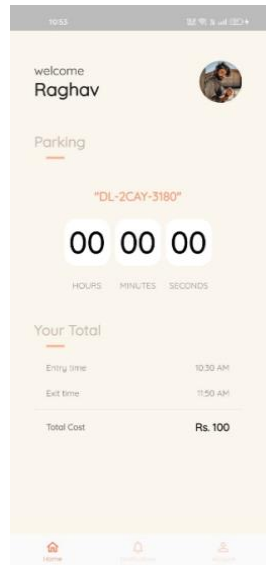

(e)

**Fig. 9.** (a) A vehicle entering the parking facility (b) License Plate detected by ANPR (c) Cropped Image from ANPR module (d) Extracted module from OCR module (e) App interface with number details and parking trip details

## 6  Conclusion

We created a microcontroller and AI-based system that can easily manage difficult parking systems in this work. To provide a unified solution, this system is paired with an excellent front-end application. We analysed and fine-tuned several object identification models, and we used the OCR process to select the best combination to use. Further research into quicker OCR pipelines and enhanced object detection models may make the system even more reliable and future-proof. Microcontrollers with faster processor speeds can reduce execution times even further. Ultimately, the construction delivers a full parking system solution.